\documentclass[letterpaper, 10 pt, conference]{ieeeconf}  

\IEEEoverridecommandlockouts                              

\usepackage{graphicx} 
\usepackage{amsmath} 
\usepackage{dsfont}
\usepackage{multirow}
\usepackage{tabularx}
\usepackage{booktabs}
\usepackage{amsfonts}       
\usepackage{mathtools, nccmath}
\usepackage{caption}
\usepackage{float}
\newlength\savewidth
\newcommand\shline{\noalign{\global\savewidth\arrayrulewidth
  \global\arrayrulewidth 1pt}\hline\noalign{\global\arrayrulewidth\savewidth}}

\title{\LARGE \bf
TrajSSL: Trajectory-Enhanced Semi-Supervised 3D Object Detection
}

\author{Philip Jacobson$^{1}$, Yichen Xie$^{1}$, Mingyu Ding$^{1}$, Chenfeng Xu$^{1}$, \\ Masayoshi Tomizuka$^{1}$, Wei Zhan$^{1}$, and Ming C. Wu$^{1}$
\thanks{Philip Jacobson is supported by the National Defense Science and Engineering Graduate (NDSEG) Fellowship. This work is supported in part by Berkeley DeepDrive.
}
\thanks{$^{1}$University of California, Berkeley
        {\tt\small \{philip\_jacobson, yichen\_xie, myding, xuchenfeng, tomizuka, wzhan\}@berkeley.edu}
        \tt\small wu@eecs.berkeley.edu}%
}

\begin{document}

\maketitle
\thispagestyle{empty}
\pagestyle{empty}

\begin{abstract}

Semi-supervised 3D object detection is a common strategy employed to circumvent the challenge of manually labeling large-scale autonomous driving perception datasets. Pseudo-labeling approaches to semi-supervised learning adopt a teacher-student framework in which machine-generated pseudo-labels on a large unlabeled dataset are used in combination with a small manually-labeled dataset for training. In this work, we address the problem of improving pseudo-label quality through leveraging long-term temporal information captured in driving scenes. More specifically, we leverage pre-trained motion-forecasting models to generate object trajectories on pseudo-labeled data to further enhance the student model training. Our approach improves pseudo-label quality in two distinct manners: first, we suppress false positive pseudo-labels through establishing consistency across multiple frames of motion forecasting outputs. Second, we compensate for false negative detections by directly inserting predicted object tracks into the pseudo-labeled scene. Experiments on the nuScenes dataset demonstrate the effectiveness of our approach, improving the performance of standard semi-supervised approaches in a variety of settings.

\end{abstract}

\begin{figure*}[thpb]
    \centering
    \includegraphics[scale=0.4]{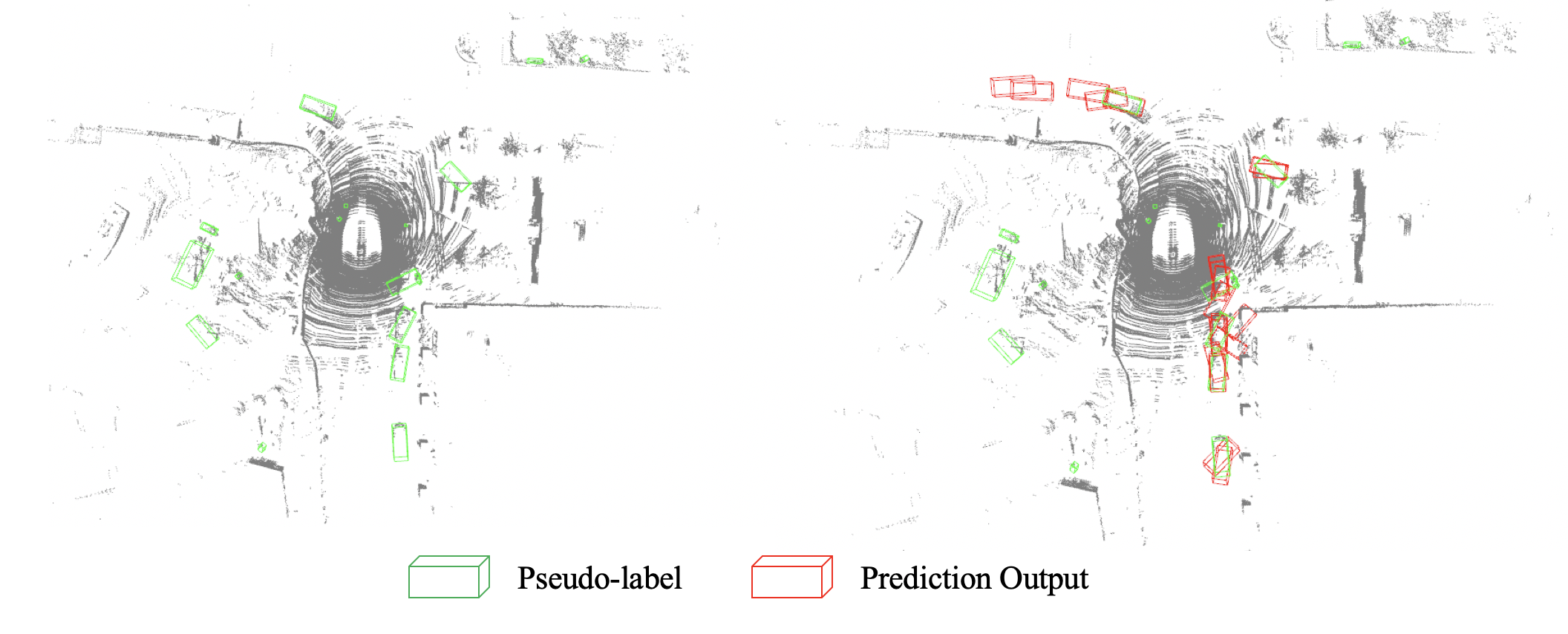}
    \caption{Comparison between a scene containing only teacher-generated pseudo-labels (in green), and the scene augmented with both pseudo-labels and predicted trajectory boxes (in red). Overlapping red and green boxes indicate pseudo-labels exhibiting a high degree of temporal consistency, which are further emphasized during student training. Green boxes without overlap indicate pseudo-labels exhibiting a low degree of temporal consistency, and hence more likely to be a false positive detection. Unmatched red boxes indicate potential missed detections by the teacher model, and are also added as soft targets during training.}
    \label{fig:predcomparison}
\end{figure*}

\section{INTRODUCTION}

3D object detection is a key task within the autonomous driving perception stack. While many LiDAR point cloud-based methods are able to achieve impressive performance \cite{yin2021center,pvrcnn++,lidarmultinet,liu2023flatformer}, training these models requires large-scale labeled point cloud datasets. In contrast to procuring labeled 2D image data, labeling 3D point clouds for object detection tasks is a niche skill set; as a result manual labeling is both expensive and time-consuming. Thus, the challenge of acquiring human-labeled 3D detection data is a significant bottleneck to training the powerful 3D object detectors needed for autonomous vehicles.

Semi-supervised learning (SSL), or the idea of learning with a small labeled dataset in combination with a large unlabeled dataset, is a popular framework for label-efficient training of machine learning models. One approach to semi-supervised learning, known as self-training or pseudo-labeling, uses a pre-trained teacher model to generate pseudo-labels on the large body of unlabeled data, before training a student model on a mixture of labeled/pseudo-labeled data. Various approaches have been proposed for applying pseudo-labeling to both 2D object detection ~\cite{liu2021unbiased,NEURIPS2019_d0f4dae8,Tang2021HumbleTT,sohn2020detection,zhou2022} and 3D object detection ~\cite{park2022detmatch,wang20213dioumatch,li2023dds3d,liu2023hierarchical,Chen_2023_WACV}. All of these works seek to address a key challenge of pseudo-labeling: what is the best strategy for maximizing supervision from high-quality pseudo-labels during training, while minimizing supervision from low-quality ones? 

In order to address this problem, we first need a quantifiable measure of pseudo-label quality. In the context of object detection, a rudimentary approach is to simply use the teacher model detection confidence score as a proxy for pseudo-label quality. However, particularly for a teacher model trained on a limited dataset, the confidence score is often weakly correlated with a pseudo-label's true agreement with a ground truth label \cite{park2022detmatch}. Other works seek to use some form of consistency measure, such as consistency between augmented views \cite{liu2023hierarchical}, consistency between differing modalities \cite{park2022detmatch}, or consistency between pseudo-labels and ground truth labels on labeled data \cite{doublyrobust} as measures of pseudo-label quality. Through establishing an improved measure of pseudo-label quality, these methods attempt to strike a careful balance between identifying likely false positive pseudo-labels, while not being so stringent as to unintentionally create new false negatives through misidentification of valid pseudo-labels.

In the autonomous driving setting in which object detection is inherently linked to navigating dynamic scenes over time, temporal sequence inputs offer an opportunity for improved detection performance. Several methods for multi-frame 3D object detection have been proposed in the literature ~\cite{chen2022mppnet,He_2023_CVPR,qi2021offboard,3DMan}. One previous work, MoDAR, leverages motion forecasting as a vehicle for propagating temporal information, generating virtual points which are added to the point cloud \cite{modar}. However, few works have explored leveraging temporal inputs in the context of semi-supervised object detection.

In this work, we propose leveraging outputs from trajectory prediction models to improve pseudo-label supervision during semi-supervised training, which we dub TrajSSL. We build our method on top of the standard teacher-student framework for SSL. First, during the teacher model pre-training stage, we additionally pre-train a trajectory prediction model on the labeled data split available to us. During teacher inference on the unlabeled data, we run a multi-object tracker to link pseudo-labels into object tracks to then be used as inputs to our pre-trained prediction model. Using our forecasting model, we generate future motion trajectories based on the tracked pseudo-labels; outputs are then assigned to the corresponding future frame, such that at the end of inference each frame in the unlabeled set contains a set of objects predicted based on varying context frames. During student training, we use these virtual objects in two differing manners. First, to identify strong pseudo-labels, we measure IoU overlap between virtual objects and pseudo-labels; as pseudo-labels overlapping predicted trajectories exhibit a degree of temporal consistency, we increase the weight of these labels in the training objective, scaled by the number of overlaps. Second, we compensate for false negative detections through inserting unmatched virtual objects into the set of pseudo-labels to add extra supervision during training. Fig. \ref{fig:predcomparison} visualizes the effect of augmenting the teacher model pseudo-labels with predicted trajectories during training.

We validate TrajSSL using the nuScenes autonomous driving dataset, as it is readily compatible with both open-source 3D detection and trajectory prediction models. Performing experiments in a wide variety of experimental settings, we demonstrate absolute improvement in mAP over previous semi-supervised 3D object detection methods.

\section{Related Work}
\subsection{3D Object Detection}
A few broad strategies exist for point cloud-based 3D object detection. Point-based methods directly ingest the point cloud ~\cite{qi2019deep,pointgnn,Shi_2019_CVPR,Yang2019std,3DSSD}, grouping points in a bottom-up manner to enable hierarchical learning with PointNet-based \cite{Qi_2017_CVPR} feature extractors. Voxel-based methods ~\cite{voxelnet,Lang2019PointPillarsFE,second,yin2021center,lidarmultinet,liu2023flatformer,Zhou_centerformer,sun2022swformer,mao2021voxel} generate a regularized voxel grid from the point cloud to enable compatibility with standard neural architectures, such as CNNs and transformers. VoxelNet \cite{voxelnet} encodes the point cloud into voxel features using a PointNet-like architecture to then be processed by a 3D CNN region proposal network. PointPillars \cite{Lang2019PointPillarsFE} operates in a similar manner, however instead discretizes the space into 2D pillars with infinite height to enable faster encoding. CenterPoint \cite{yin2021center} adopts a voxel-based backbone while performing detection with an anchor-free approach. Transformer-based approaches such as SWFormer \cite{sun2022swformer} and Flatformer \cite{liu2023flatformer} replace the 3D CNN backbone with shifted-window transformers. PV-RCNN ~\cite{pvrcnn++,Shi_2020_CVPR} uses a hybrid point-voxel approach to leverage the benefits of both types of feature extraction. Multi-frame object detectors such as MPPNet \cite{chen2022mppnet} and 3DAL \cite{qi2021offboard} use a two-stage refinement where inputs from multiple frames are used to improve bounding box estimates. 

\subsection{Trajectory Prediction}
Decision-making in robots/autonomous vehicles navigating dynamic scenes requires an awareness of the motion of other agents in the scene. Trajectory prediction uses the historical motion of other agents in combination with scene-level information (e.g. HD maps) to forecast future agent trajectories. A variety of approaches to exist to trajectory prediction ~\cite{yuan2021agent,Salzmann2020TrajectronDT,multipath,liu2021multimodal,deo2021multimodal}, generally relying on neural generative modeling to produce future object trajectories. Agentformer \cite{yuan2021agent} jointly models both temporal and social interactions between agents in the scene, generating trajectories using a conditional variational autoencoder (CVAE) generative model. A few works have also examined training prediction models in a label-efficient manner \cite{xu2022pretram,chen2023unsupervised}, although this direction remains generally unexplored.

\subsection{Semi-supervised Object Detection}
Initial works on semi-supervised object detection primarily focused on the 2D detection task ~\cite{liu2021unbiased,NEURIPS2019_d0f4dae8,Tang2021HumbleTT,sohn2020detection,zhou2022,instantzhou2021}. STAC \cite{sohn2020detection} strongly augments inputs to the student model to enforce augmentation consistency between pseudo-labels. Unbiased teacher \cite{liu2021unbiased} uses an exponential moving average (EMA) to update the teacher model during student training. More recent works have also investigated semi-supervised 3D object detection ~\cite{park2022detmatch,wang20213dioumatch,li2023dds3d,liu2023hierarchical,Chen_2023_WACV,doublyrobust,Zhao_2020_CVPR}. SESS \cite{Zhao_2020_CVPR} utilizes three consistency losses to enforce agreement between perturbed variations of the input data. 3DIoUMatch \cite{wang20213dioumatch} utilizes an IoU estimation module score as a confidence threshold filter. DetMatch \cite{park2022detmatch} takes a multi-modal approach, using agreement between camera model pseudo-labels and LiDAR model pseudo-labels to filter pseudo-labels. HSSDA \cite{liu2023hierarchical} uses an improved strong data augmentation scheme in combination with hierarchical supervision based on pseudo-label quality to improve training. Playbacks for UDA \cite{you2022}, similar to our work, also adopts a temporal refinement of pseudo-labels, using a tracking interpolation/extrapolation module to improve pseudo-label quality in the context of unsupervised domain adaptation.

\section{Method}
\begin{figure*}[thpb]
    \centering
    \includegraphics[scale=0.4]{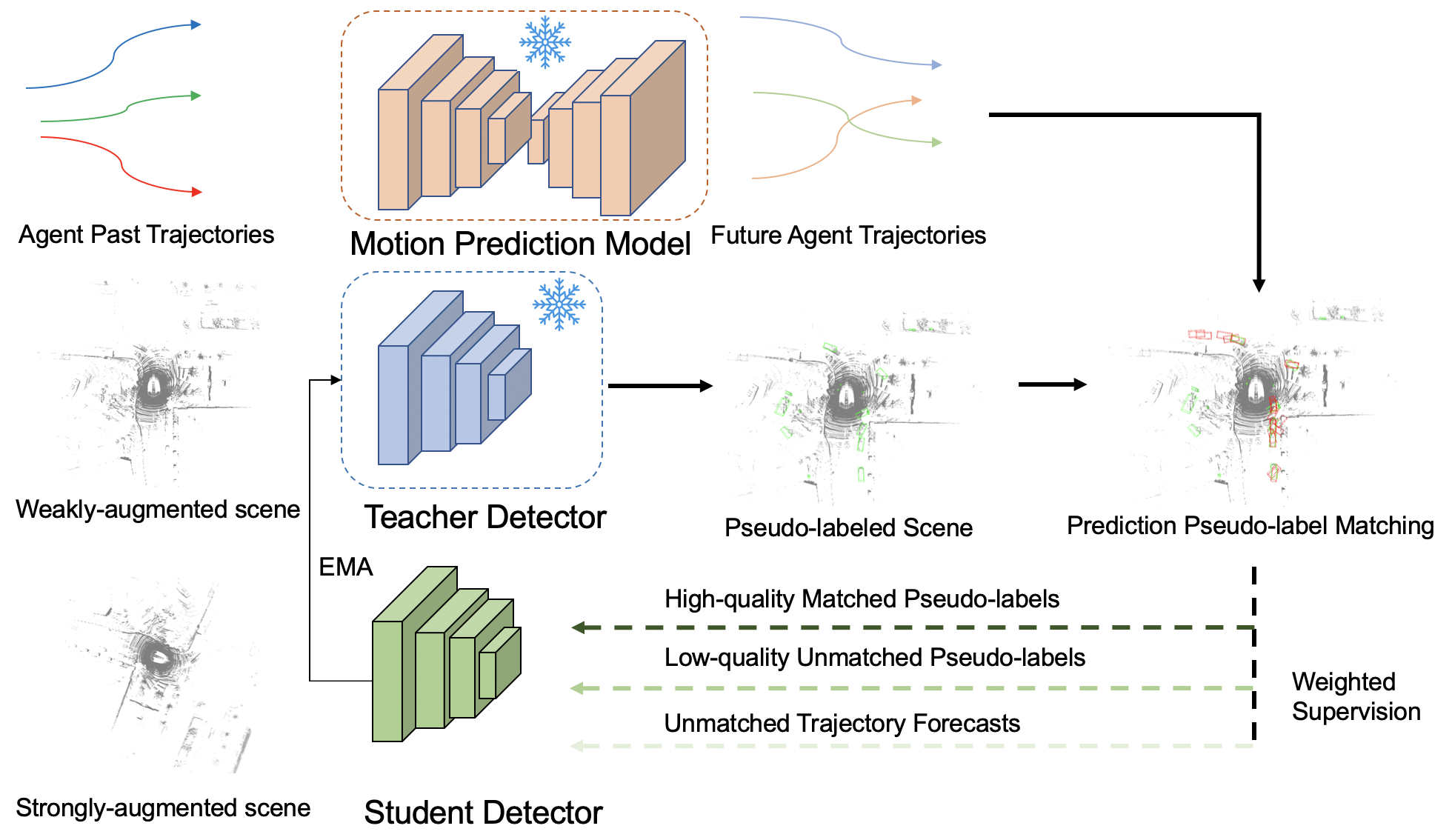}
    \caption{Overview of our proposed method TrajSSL. In addition to a teacher-student SSL framework, we introduce a trajectory prediction model (AgentFormer) which predicts future object trajectories based on past pseudo-label tracks. The inference output of this model is combined with the perception pseudo-labels and an IoU=matching process is performed. Pseudo-labels are then weighted during supervision based on the degree to which they agree with the forecasted trajectories. Meanwhile, predictions which don't match already existing pseudo-labels are added to the training process as down-weighted pseudo-labels.}
    \label{fig:trajssl}
\end{figure*}

In this section, we introduce our proposed approach TrajSSL, and describe in detail both the generation of synthetic trajectories, and the semi-supervised training of a student model leveraging these trajectory outputs. An overview of our approach is shown in Fig. \ref{fig:trajssl}.

\subsection{Problem Definition}
In the semi-supervised setting, we have at our disposal two sets of data: a set of manually annotated samples $\mathcal{D}_l = \{(\textbf{x}_i^l,\textbf{y}_i^l )\}_{i=1}^{N_{l}}$, and a set of unlabeled samples $\mathcal{D}_u = \{\textbf{x}_i\}_{i=1}^{N_u}$. Typically we are only able to annotate a small fraction of our data, meaning $N_u >> N_l$. For point-cloud based 3D object detection, our input data samples consist of a list of unordered points $\mathcal{P} = \{(x_i,y_i,z_i,r_i)\}$, where $(x,y,z)$ denote the Cartesian 3D coordinate and $r$ denotes the reflectance measured by the LiDAR sensor. Each sample label consists of a set of bounding boxes $\mathcal{B} = {b_i}$, with each box $b$ consisting of a class description and 7 localization parameters: center 3D location, box size, and box orientation.

\subsection{Teacher-Student Framework}
TrajSSL is built on the frequently-used teacher-student paradigm of SSL. For our experiments, we employ a CenterPoint \cite{yin2021center} with PointPillars \cite{Lang2019PointPillarsFE} backbone as our detector models, however any off-the-shelf 3D detector is compatible with this paradigm. First, the teacher model $\textbf{T}$ is pre-trained on the labeled data samples $\mathcal{D}_l$ until convergence. During student training, the teacher model performs inference on the unlabeled dataset to generate pseudo-labels. The student model $\textbf{S}$ is then trained on the combination of labeled samples $\{(\textbf{x}^l_i,\textbf{y}^l_i)\}_i$ and pseudo-labeled samples $\{(\textbf{x}^u_i,\textbf{T}(\textbf{x}^u_i))\}_i$. During student model training, the teacher detector is improved using an EMA:
\begin{equation}
    \theta_\textbf{T} = \alpha \theta_\textbf{T} + (1-\alpha)\theta_\textbf{S}
\end{equation}
where $\alpha$ is the EMA momentum and $\theta_\textbf{T}$, $\theta_\textbf{S}$ are the teacher and student model parameters, respectively.

\subsection{Trajectory Generation}
\begin{figure*}[thpb]
    \centering
    \includegraphics[scale=0.35]{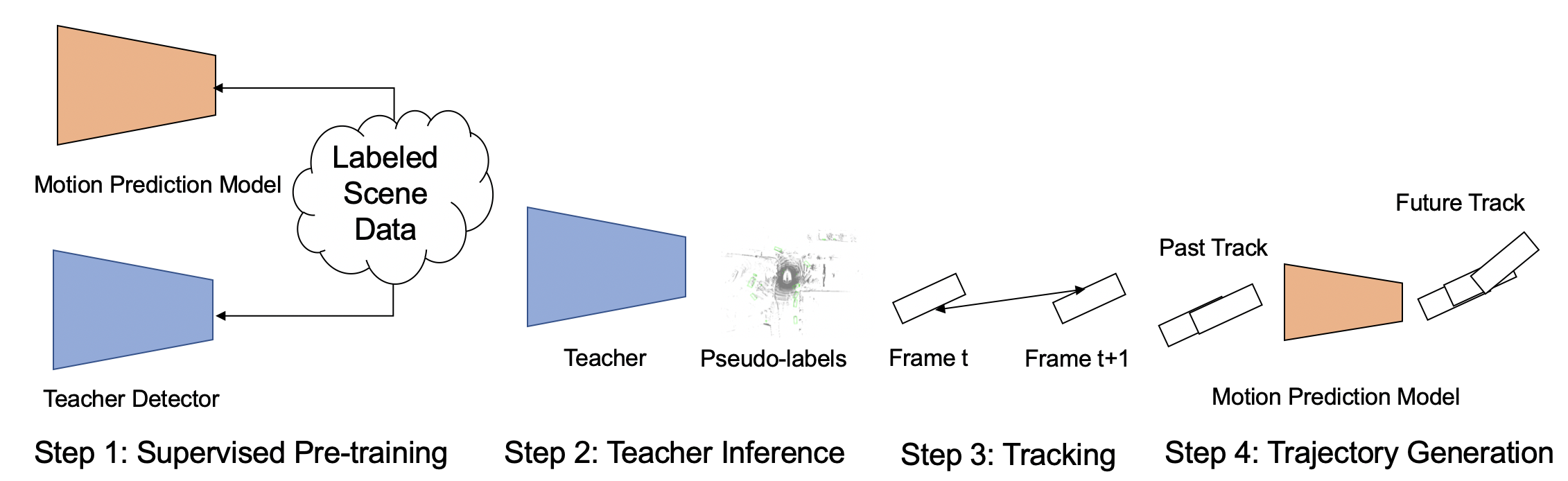}
    \caption{Illustrated process of generation trajectories from pseudo-labels. First, we pre-train both our teacher detector model and our trajectory prediction model using the available labeled scene data. Next, we use the teacher model to run inference on the unlabeled scene data. Next, we link the produced pseudo-labels into tracks of objects across time. Lastly, we feed these tracks into prediction model to generate synthetic trajectories.}
    \label{fig:trainingflow}
\end{figure*}

During the teacher pre-training stage, we additionally pre-train a trajectory prediction model for use in the downstream training. For our work, we adopt Agentformer \cite{yuan2021agent} as our motion forecasting model of choice, although our method is compatible with any off-the-shelf model. Agentformer takes two sets of inputs: a set of agent histories, $\{ (\textbf{x}_i^{-H},\textbf{x}_i^{-H+1},...,\textbf{x}_i^0\}_{i=1}^N$ for up to $H+1$ timesteps, and optionally an HD scene-level semantic map. As output, Agentformer generates a set of future trajectory predictions for each input agent, $\{ (\textbf{p}_i^{1},\textbf{p}_i^{2},...,\textbf{p}_i^T\}_{i=1}^N$ for up to $T$ future timesteps. In this initial stage, Agentformer is pre-trained using the same labeled data split available for semi-supervised training. After completing the pre-training stage, we run teacher model inference on the unlabeled dataset, followed by a multi-object tracker, to generate linked pseudo-label tracks to be used as inputs to Agentformer. Next, we run trajectory prediction inference on all frames of pseudo-labeled scenes, grouping prediction outputs according to their timestamp. Thus, for a sample in the unlabeled set with scene timestamp $t$, it now has a set of predicted agent locations grouped by prediction context frames: $\{\textbf{p}_i^{t-T},\textbf{p}_i^{t-T+1},...,\textbf{p}_i^{t-1}\}$. A summary of this process is shown in Fig. \ref{fig:trainingflow}.

\subsection{Matched Prediction Pseudo-label Weighting}
After trajectory generation, we now have a set of additional labels to aid in the training of the student detector in addition to the teacher-generated pseudo-labels. The first key insight we exploit is using object forecasts as a measure of \textit{temporal consistency}. If our prediction model predicts a consistent localization for an agent in the scene at a given future timestamp for differing input temporal frames, we argue that this hallucinated object exhibits a strong temporal consistency. Furthermore, if a pseudo-label overlaps with one of these forecasted objects, we can deduce it is likely a higher-quality label, and less likely to be a false positive detection. Thus, by computing the overlap between pseudo-labels and prediction boxes, we have an effective metric for suppressing spurious detections, and emphasizing high-quality labels. To do so, we first compute a maximum IoU between the pseudo-labels and each set of grouped prediction outputs, grouped by context frame. We set a threshold $\tau_{min\_iou}$ to use for determining whether a pseudo-label and prediction output are successfully ``matched". Then, we calculate a per pseudo-label weight based on the number of overlaps meeting the IoU threshold. For the $i^{th}$ pseudo-label, we express this quantitatively as:
\begin{equation}
\label{eq:weight}
    w_i = \alpha + \sum_{j=t-T}^{t-1} \beta \mathds{1}\{\max(IoU(\textbf{x}_i,\{\textbf{p}|\textbf{p}\in \textbf{p}^j\})) \geq \tau_{min\_iou}\}
\end{equation}
where $\mathds{1}$ is the indicator function and  $\alpha$ and $\beta$ are hyperparameters. The upshot of this weighting scheme is a linear scale for which a greater number of overlapping prediction outputs generates a higher weight. These weights are then used during pseudo-label supervised learning, explained in Sec \ref{sec:loss}.

\subsection{Unmatched Prediction-Enhanced Training}
While our pseudo-label prediction matching module acts as a filter for pseudo-labels, we also want to be able to correct for the other main source of pseudo-label inaccuracies: false negative (i.e. missed) detections. Our second key insight is in regards to \textit{unmatched} prediction outputs; we note that objects that are missed detections by the teacher model in the current frame, but are successfully tracked in any preceding frames can be recovered based on the forecasted trajectory. Therefore, we propose directly inserting unmatched prediction outputs into the pseudo-label set used during training. To determine unmatched predictions, we once again calculate the maximum IoU between each prediction box and the pseudo-label set. We set a threshold $\tau_{max\_iou}$, which is used as the maximum IoU any prediction box can have with a pseudo-label and still be considered ``unmatched". We note that in general $\tau_{max\_iou} \neq \tau_{min\_iou}$. While we can directly treat each unmatched detection in a manner equal to a teacher model detection, objects generated by the motion forecasting model are also affected by inaccuracies inherent to predicting future scenes, and thus should not be treated as equivalent to a perceived object. Instead we generate a set of linearly decreasing weights $\gamma_{t-1}, \gamma_{t-2}, ..., \gamma_{t-T}$, where $\gamma_{t-1} \leq 1$, corresponding to a given prediction context frame. We then add each unmatched prediction and assign it the $\gamma$ value corresponding to the context frame used to generate it. Since our trajectory prediction model becomes less accurate the further in the future it forecasts, we weight unmatched predictions from more recent context frames with greater weight than predictions from further in the past.

\subsection{Training Objective}
\label{sec:loss}

During semi-supervised training, we freeze the teacher model weights and only train the student model. We supervise the student model $\textbf{S}$ with two loss functions: $\mathcal{L}_l$ and $\mathcal{L}_u$, corresponding to the loss on unlabeled and labeled data, respectively.

\begin{equation}
    \mathcal{L}_l = \sum_i \mathcal{L}_{reg}(\textbf{S}(\textbf{x}_i^l),\textbf{y}_i^l) + \mathcal{L}_{cls}(\textbf{S}(\textbf{x}_i^l),\textbf{y}_i^l) 
\end{equation}

\begin{align}
\begin{split}
    \mathcal{L}_u = \sum_i \biggl( \sum_j  w_{ij}\mathcal{L}_{reg}(\textbf{S}(\textbf{x}_i^u)_j,\textbf{T}(\textbf{x}_i^u)_j) + w_{ij}  \mathcal{L}_{cls}(\textbf{S}(\textbf{x}_i^u)_j,
    \\
    \textbf{T}(\textbf{x}_i^u)_j)     
    +\sum_k w_{ik}\mathcal{L}_{reg}(\textbf{S}(\textbf{x}_i^u)_k,\Tilde{p}_{ik})
    \\
    + w_{ik}\mathcal{L}_{cls}(\textbf{S}(\textbf{x}_i^u)_k,\Tilde{p}_{ik}) \biggr)
\end{split}
\end{align}
where $\mathcal{L}_{cls}$ is the classification loss, $\mathcal{L}_{reg}$ is the bounding box regression loss, $w_{ij}$ is the weight corresponding to the $j^{th}$ pseudo-label of the $i^{th}$ frame, and $\Tilde{p}_{ik}$ is the $k^{th}$ unmatched prediction output of the $i^{th}$ frame. During training, we enforce a 1:1 batch ratio of labeled scenes to unlabeled scenes. Thus, the total training objective is defined as simply the sum of the two losses:
\begin{equation}
    \mathcal{L}_{tot} = \mathcal{L}_u + \mathcal{L}_l
\end{equation}

\section{Experiments}
To validate our approach, we perform experiments on the nuScenes dataset, a large-scale autonomous driving dataset \cite{nuscenes2019}. nuScenes consists of 1000 annotated 20-second driving scenes (700 training, 150 validation, and 150 test). In addition to LiDAR point clouds, camera images and radar point clouds, scene-level HD semantic maps are provided as data inputs. The main detection metrics used for the nuScenes object detection task are mean-average precision (mAP) and the nuScenes detection score (NDS), a dataset-specific custom metric consisting of an average of mAP and five false-positive metrics. Although nuScenes object labels are broken down into ten classes, we restrict our evaluation to the three classes compatible with Agentformer's released models: trucks, cars, and busses. For a comparison baseline, we adopt unbiased teacher \cite{liu2021unbiased} with a tuned confidence threshold filtering, which we denote as ``confidence thresholding", as similarly proposed in \cite{park2022detmatch}.

\subsection{Implementation Details}
We implement our approach using Centerpoint PointPillars as the detection backbones, and Agentformer as our trajectory prediction model. During the pre-training stage, we pre-train both the teacher detection model and Agentformer on the same split of labeled nuScenes training data. For pre-training the detection model, we follow the standard nuScenes training setting outlined in \cite{zhu2019class}, while for pre-training Agentformer we follow the training scheme used in the official implementation \cite{yuan2021agent}.

After running teacher model inference on the unlabeled data, we first filter the extracted pseudo-labels with a detection confidence of $\tau_{conf} = 0.3$. To link the extracted pseudo-labels into tracks, we use the greedy tracking algorithm used in \cite{yin2021center}. When running AgentFormer inference, we forecast trajectories only for tracks containing at least two frames of past context, while allowing for up to four frames of input. AgentFormer produces up to 12 future frames of trajectory data, and we extract predictions on all scene frames for which there is at least a single future frame in the dataset. As AgentFormer only predicts the $(x,y)$ location of an agent in BEV space, we assign the other bounding box attributes of the predicted object according to the attributes of the pseudo-label in the present context frame. 

\begin{table*}[h]
\centering
\small
\vspace{0.15in}
\begin{tabular}{c||c c c| c c c| c c c }
\shline
\multirow{2}{*}{Method} & \multicolumn{3}{c|}{5\%} & \multicolumn{3}{c|}{10\%} & \multicolumn{3}{c}{20\%} \\
     & car & truck & bus & car & truck & bus & car & truck & bus \\
 \shline
 Labeled Only & 49.1 & 8.7 & 3.2 & 61.0 & 14.2 & 8.6 & 66.9 & 23.0 & 22.5\\
 \hline
 SSL Baseline*  & 52.9 & 11.2 & 4.6 & 63.2 & \textbf{15.8} & 9.9 & \textbf{70.9} & 24.4 & 27.0\\
 Improvement & +3.8 & +2.5 & +1.4 & +2.2 & +1.6 & +1.3 & +4.0 & +1.4 & +4.5\\
 \hline
 Doubly Robust Training* & 53.7 & 11.0 &  5.9 & 64.1 & 14.7 & 11.0 & \textbf{70.9} & 24.3 & 26.4   \\
 Improvement & +4.6 & +2.3 & +2.7 & +3.1 & + 0.5 & +2.4 & +4.0 & +1.3 & +3.9  \\
 \hline
 Ours & \textbf{54.3} & \textbf{11.4} & \textbf{9.3} & \textbf{64.7} & 15.7 & \textbf{11.9} & 70.1 & \textbf{24.8} & \textbf{27.5} \\
 Improvement & +5.2 & +2.7 & +6.1 & +3.7 & +1.5 & +3.3 & +3.2 & +1.8 & +5.0\\
 \shline
\end{tabular}
\caption{Performance (mAP) comparison on nuScenes validation dataset for car, truck and bus class on a variety of labeled data fraction settings. Our proposed TrajSSL improves performance over previous semi-supervised approaches across all classes in a wide variety of settings. *our re-implementation}
\label{tab:mainresults}
\end{table*}

\subsection{Main Results}
We evaluate TrajSSL on the nuScenes dataset for three different labeled data settings: training with 5\% labeled data, 10\% labeled data, and 20\% labeled data. We summarize these results in Tab. \ref{tab:mainresults}. Across all three settings, TrajSSL improves performance over the confidence thresholding baseline with generally strong performance for all three classes. In the setting with the least labeled data available, we see the most significant performance gains from TrajSSL; in particular, the car and bus classes see an improvement of 1.4 and 4.7 mAP points over the baseline. As the labeled data available increases and the teacher model becomes stronger (hence there exists fewer false positives/negatives to correct for), the relative improvement gained by TrajSSL decreases, though is still noticeable. Additionally, we also compare our approach to doubly-robust training \cite{doublyrobust}, a more general SSL framework. Across all settings and classes, TrajSSL outperforms doubly-robust training. Notably, in the 20\% labeled data setting, in which doubly-robust training fails to improve over the confidence thresholding baseline, TrajSSL is still able to gain modest improvements in the bus and truck classes.

\subsection{Ablation Studies}
In this section, we perform ablation studies on the various aspects of our TrajSSL framework. We perform all ablation experiments using the 5\% labeled training data setting.

\textbf{False Positive/Negative Compensation.} The first set of ablation experiments we perform is to verify the improvement gained from our two strategies for suppressing false positives and directly correcting false negatives. We summarize the results of these experiments in Tab. \ref{tab:mainablation}. We find the most significant improvement arises from the up-weighting of pseudo-labels which are matched to a prediction output; while the improvement to the truck class is modest, the bus and car class see an improvement of +4.3 mAP and +1.2 mAP, respectively. This supports our hypothesis of temporal consistency established through trajectory forecasts being a good metric for pseudo-label quality. 

Our second key component, direct addition of prediction outputs to correct false negatives, results in a further modest increase in performance, improving the car and bus class by +0.2 mAP and +0.4 mAP, respectively while truck class mAP remains unchanged. While the ability to directly replace false negatives with forecasted objects is limited by the quality of the pseudo-label tracks used as input to Agentformer, nonetheless a consistent improvement verifies that unmatched prediction objects contain useful information gained from temporal context and can improve the student model training.

\begin{table}[h]
\centering
\small
\resizebox{\linewidth}{!}{
\begin{tabular}{c||c | c | c }
\shline
 & Car & Truck & Bus\\
 \shline
 Labeled Only & 49.1 & 8.7 & 3.2 \\
 \hline
 + Teacher-Student SSL & 52.9 & 11.2 & 4.6 \\
 \hline
 + Matched Prediction Pseudo-label Weighting & 54.1 & 11.4 &  8.9 \\
 \hline
 + Unmatched Prediction Addition & \textbf{54.3} & \textbf{11.4} & \textbf{9.3}\\
 \shline
\end{tabular}}
\caption{Ablation of two main strategies of TrajSSL.}
\label{tab:mainablation}
\end{table}

\textbf{Trajectory Time Horizon.} The next key aspect of our approach we want to verify is the utility of Agentformer's future predictions. To do so, we perform experiments using a varying number of temporal frame outputs from Agentformer, which is capable of predicting up to 12 frames (6 seconds in the context of nuScenes) into the future. We include the results in Tab. \ref{tab:frameablation}. We see that adopting TrajSSL for even one single frame of trajectory outputs significantly improves performance over the non-temporal baseline. Increasing the number of Agentformer output frames to 5 frames results in a further increase in mAP, although the improvement is far less dramatic then the jump from one to two frames. Going further to 8 or 10 frames degrades performance from using 5 frames for both the car and bus class, while slightly improving the truck class by +0.1 mAP, indicating forecasted objects this far into the future aren't accurate enough to successfully integrate into TrajSSL.

\begin{table}[h]
\centering
\small
\begin{tabular}{c||c | c | c }
\shline
 & Car & Truck & Bus\\
 \shline
 +1 Frame (SSL Baseline) & 52.9 & 11.2 & 4.6 \\
 \hline
 +2 Frames & 53.9 & 11.0 & 8.5 \\
 \hline
 +5 Frames & \textbf{54.3} & 11.4 & \textbf{9.3} \\
 \hline
 +8 Frames & 53.8 & \textbf{11.5} & 8.7\\
 \hline
 +10 Frames & 53.9 & \textbf{11.5} & 8.8\\
 \shline
\end{tabular}
\caption{Ablation of number of prediction frames used in TrajSSL.}
\label{tab:frameablation}
\end{table}

\textbf{Linear Extrapolation Baseline Comparison.} A further ablation study we perform is to directly probe the necessity of a complex neural model for generating the future forecasts of scene objects. As a baseline, we consider performing a linear extrapolation using the model-predicted velocity of each object to predict future object locations, after which we use our already proposed weighting mechanism. We compare these two approaches in Tab. \ref{tab:linearextrap}. Using the linear extrapolation approach is still able to improve the SSL baseline on both the car and bus class. However, across all three classes, predicting future trajectories using Agentformer noticeably outperforms the simple linear extrapolation approach. We attribute this to the fact that a) the teacher model (particularly when pre-trained on limited data) is poor at predicting velocity accurately, making linear extrapolation less accurate and b) particularly for longer time-horizon forecasting, linear extrapolation is too simple to capture the complex scene dynamics to accurately predict agent trajectories. Thus, a powerful trajectory prediction model, even when trained on a sparse dataset, is a key ingredient to maximizing the effectiveness of TrajSSL.

\begin{table}[h]
\centering
\small
\begin{tabular}{c||c | c | c }
\shline
 & Car & Truck & Bus\\
 \shline
 SSL Baseline & 52.9 & 11.2 & 4.6 \\
 \hline
 Prediction Model (AgentFormer) & \textbf{54.3} & \textbf{11.4} & \textbf{9.3}\\
 \hline
 Linear Extrapolation & 53.2 & 11.0 & 8.4 \\
 \shline
\end{tabular}
\caption{Comparison of our approach using Agentformer versus using a linear extrapolation.}
\label{tab:linearextrap}
\end{table}

\section{Conclusion}
In this paper, we proposed a novel framework for semi-supervised 3D object detection in autonomous driving scenarios based on leveraging trajectory prediction models to enhance pseudo-label training, which we dub TrajSSL. TrajSSL uses outputs from Agentformer, a trajectory forecasting model, to enhance the training of the student detector in two key ways: first, it uses these predicted objects to locate higher-quality pseudo-labels and up-weight them during the training process. Second, unmatched outputs are used to directly compensate for missed detections. On experiments using the nuScenes dataset, TrajSSL outperforms previous SSL approaches in a wide variety of settings.

\bibliographystyle{IEEEtran}
\bibliography{ref}

\end{document}